\documentclass[runningheads]{llncs}
\usepackage{graphicx}
\usepackage{filecontents}
\usepackage{amsmath,amssymb} 
\usepackage{color}

\begin{document}
\title{U-Finger: Multi-Scale Dilated Convolutional Network for Fingerprint Image Denoising and Inpainting} 

\titlerunning{Fingerprint Denoising}
%
\author{Ramakrishna Prabhu\inst{1} \and
Xiaojing Yu\inst{1} \and
Zhangyang Wang\inst{1} \and Ding Liu\inst{2} \and Anxiao (Andrew) Jiang\inst{1}}
%
\authorrunning{ }
%

\institute{ Texas A\&M University, College Station, TX 77843, USA
\email{\{rgsl888,xiaojingyu,atlaswang,ajiang\}@tamu.edu}\\
\and
University of Illinois at Urbana-Champaign, IL, USA\\
\email{dingliu2@illinois.edu}}
\maketitle              
\begin{abstract}
This paper studies the challenging problem of fingerprint image denoising and inpainting. To tackle the challenge of suppressing complicated artifacts (blur, brightness, contrast, elastic transformation, occlusion, scratch, resolution, rotation, and so on) while preserving fine textures, we develop a multi-scale convolutional network termed \textit{U-Finger}. Based on the domain expertise, we show that the usage of dilated convolutions as well as the removal of padding have important positive impacts on the final restoration performance, in addition to multi-scale cascaded feature modules. Our model achieves the overall ranking of No.2 in the \textit{ECCV 2018 Chalearn LAP Inpainting Competition Track 3} (Fingerprint Denoising and Inpainting). Among all participating teams, we obtain the MSE of 0.0231 (rank 2), PSNR 16.9688 dB (rank 2), and SSIM 0.8093 (rank 3) on the hold-out testing set.

\end{abstract}
\section{Introduction}
Fingerprints are widely adopted biometric patterns in forensics applications, thanks to the growing prevalence of fingerprint sensors in daily life. The retrieval and verification of fingerprints collected in the wild, however, are often negatively impacted by poor image quality. For example, the quality of fingerprint images can easily get degraded by skin dryness, wetness, wound and other types of noise. As another example, the failure of sensors will also introduce unclear or missing local fingerprint regions, which calls for completion/inpainting. In order to ease verification carried out either by humans or existing third party software, a preprocessing step of fingerprint image restoration and enhancement has shown its practical necessity. 

Convolutional neural networks (CNNs) \cite{krizhevsky2012imagenet} have brought unprecedented success in many computer vision tasks, including some recent works addressing fingerprint extraction and analysis \cite{nguyen2018robust,tang2017fingernet}. On the other hand, fingerprint restoration and enhancement have been traditionally studied using classical example-based and regression methods \cite{singh2015fingerprint,schuch2017minutia,rahmes2007fingerprint,wang2015learning}. With lots of success of deep learning-based natural image denoising/inpainting/super resolution \cite{liu2017robust,wang2016studying,liu2017enhance,xie2012image,Liu}, the recent ECCV 2018 ChaLearn competition\footnote{\url{http://chalearnlap.cvc.uab.es/dataset/32/description/}} has started to motivate researchers to develop deep learning algorithms that can restore fingerprint images that contain artifacts such as noise, scratches \cite{wang2016d3,li2013detection}, etc. to improve the performance of subsequent operations like fingerprint verification that are typically applied to such images.

\section{Dataset and Technical Challenges}
Our work is based on the large-scale synthesized dataset of realistic artificial fingerprints, released by the ECCV 2018 ChaLearn competition organizers. They generated fingerprint images by first creating synthetic fingerprints, then degrading them with a (unknown) distortion model that will in general introduce multi-fold artifacts. They then overplayed the degraded fingerprints on top of various backgrounds to simulate the practical locations (such as walls, skins) to find fingerprints. The resulting images are typical of what law enforcement agents have to deal with. They provided a training set consisting of such original and distorted image pairs. The developed algorithms are to be tested on another non-overlapping set of distorted images, whose results will be compared against the hold-out set of corresponding original images. Developed algorithms will be evaluated based on reconstruction performance (PSNR and SSIM).

As a specific instance of learning-based image restoration, we point out a few challenges that the fingerprint image restoration task is uniquely faced with:
\begin{itemize}
	\item \textit{Complicated mixed degradation types:} the distorted images are original images added with (unknown amounts of) blur, brightness, contrast, elastic transformation, occlusion, scratch, resolution, rotation, and so on. The desired model should thus have the general ability to overcome such complicated, mixed artifacts. 
	\item \textit{Preserving fine-scale textures:} different from natural images, fingerprint images are composed of usually thin textures and edges, and it is critical to preserve and keep them sharp during the restoration process for their reliable recognition/verification from those patterns. 
\end{itemize}

\section {The Proposed Model: U-Finger}

Our proposed fingerprint denoising and inpainting model, termed \textit{U-Finger}, adopts the \textit{U}-shaped deep denoiser proposed by \cite{Liu}. As illustrated in Fig. \ref{fig1} (a), starting from the input noisy image, the network first passes the input through a few cascaded feature encoding modules followed by the same number of feature decoding modules. The output is optimized with the mean-square-error (MSE) loss with regard to the ground-truth. This network conducts feature contraction and expansion through downsampling and upsampling operations, respectively. Each pair of downsampling and upsampling operations brings the feature representation to a
new spatial scale, so that the whole network can process information at different scales.  We chose three scales in this work. Such pairs of downsampling and upsampling steps can be nested to build deeper networks, which, however, we did not do because the fine-scale features in fingerprints cannot afford convolutions that are too deep. The rich set of skip connections also helps information be passed through and avoids severe loss when the network grows deeper.

\subsubsection{Feature Encoding:} A feature encoding module is one convolutional layer plus one residual block, as displayed in Fig. \ref{fig1}(b). Note that each convolutional layer is followed by spatial batch normalization and a ReLU neuron. From top to down, the four convolutional layers have 128, 32, 32 and 128 kernels of size 3 $\times$ 3, 1 $\times$ 1, 3 $\times$ 3 and 1 $\times$ 1, respectively. 

\subsubsection{Feature Decoding:} 
\begin{figure}[tbh]
	\includegraphics[width=\textwidth]{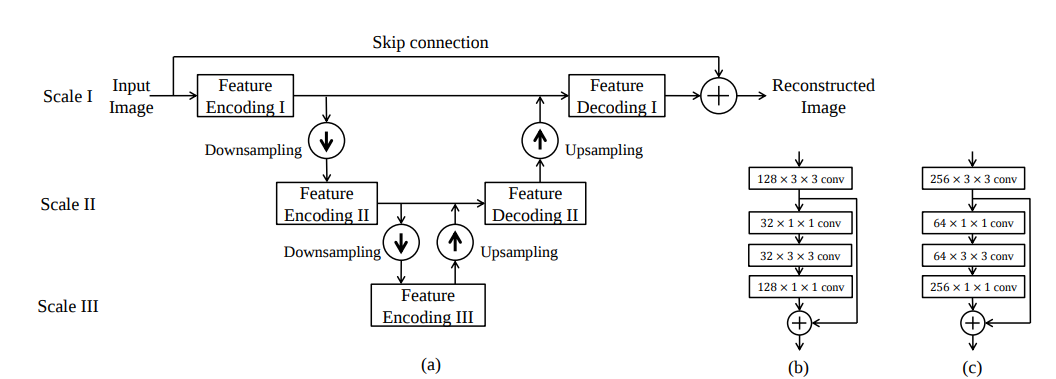}
	\caption{(a) Overview of our adopted network. (b) Architecture of the feature encoding module. (c) Architecture of the feature decoding module.}
	\label{fig1}     
\end{figure}

A feature decoding module is designed for fusing information from two adjacent scales. Two
fusion schemes are tested: (1) concatenation of features from these two scales; (2) element-wise sum of them. Both were observed to obtain similar denosing performance, and the first one was chosen by default. We use a similar
architecture as the feature encoding module except that the number of kernels in the four convolutional layers are 256,
64, 64 and 256, as in Fig. \ref{fig1}(c).

In our work, we generalize the original denosing framework in \cite{Liu} to handle more complicated mixed types of degradation. Notably, we replace the ordinary convolutions used in \cite{Liu} with the dilated convolutions \cite{yu2015multi}, with a dilation factor of 3. The dilated convolution was designed to systematically aggregate multi-scale contextual information without losing resolution, and we find it particularly better than ordinary convolutions in preserving fine fingerprint textural information throughout the deep network. We also observe that the padding adopted by the original model \cite{Liu} had a negative effect to pass on the background noise in input noisy images to the output results. While that effect was less noticeable for natural images, it can degrade the (structurally simper) fingerprint image quality quite obviously. We thus removed all paddings in our network.

\section{Experimental Results}
\begin{figure}[tbh]
	\includegraphics[width=\textwidth]{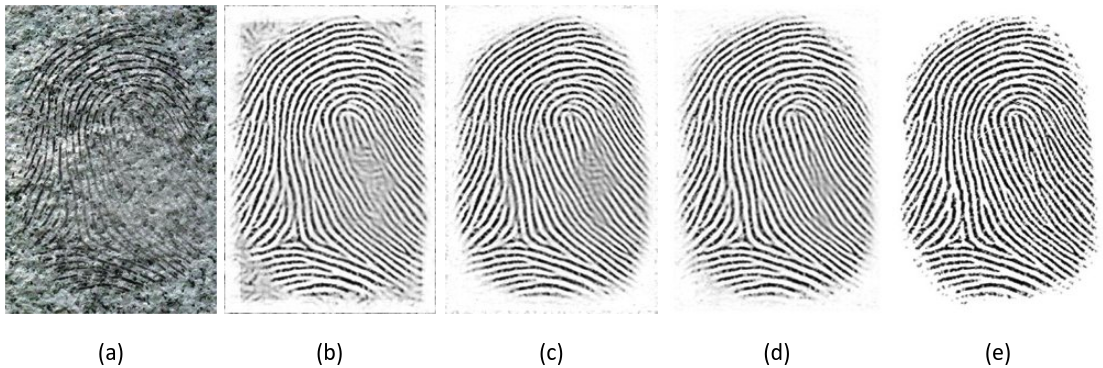}
	\caption{(a) Original, (b) Base-model (c) Base-model with no padding (d) U-Finger (e) Ground truth.}
	\label{fig2}       
\end{figure}

We train our model for 1,500,000 iterations using the stochastic gradient descent (SGD) solver with the batch size of 8. The input images are converted into gray scale. The code has been publicly released\footnote{\url{https://github.com/rgsl888/U-Finger-A-Fingerprint-Denosing-Network}}. We compare the proposed U-Finger model with the original base model in \cite{Liu}, as well as the base model with padding removed, to show the progressive performance improvements obtained. As can be seen from Table \ref{tab1}, removing padding has reduced the reconstruction errors considerably on the validation set, and the final U-Finger model with dilated convolutions has further-improved performance. Visual result examples are also displayed in Fig. \ref{fig2}, while we observe that removing padding suppresses most unwanted background artifacts without hurting fingerprint textures (since the fingerprint is pre-known to locate in the central region), and dilated convolutions are found to help preserve shaper and more consistent curvatures. 

Our model achieves the overall ranking of No.2 in the \textit{ECCV 2018 Chalearn LAP Inpainting Competition Track 3} (Fingerprint Denoising and Inpainting). Among all participating teams, we obtain the MSE of 0.0231 (rank 2), PSNR 16.9688 dB (rank 2), and SSIM 0.8093 (rank 3) on the hold-out testing set.

\begin{table}
	\centering
	\caption{MSE, PSNR and SSIM Results on Validation Set by ChaLearn Challenge Track 3}\label{tab1}
	\begin{tabular}{|l|l|l|l|}
		\hline
		&  MSE & PSNR & SSIM\\
		\hline
		Base-model & 0.029734  & 15.8747 & 0.77016\\
		Base-model without padding & 0.025813 & 16.4782 & 0.78892\\
		\bfseries U-Finger & \bfseries 0.023579 & \bfseries 16.8623 & \bfseries 0.80400\\
		\hline
	\end{tabular}
\end{table}

%

\begin{figure}[!tbh]
	\centering
	\includegraphics[width=95mm]{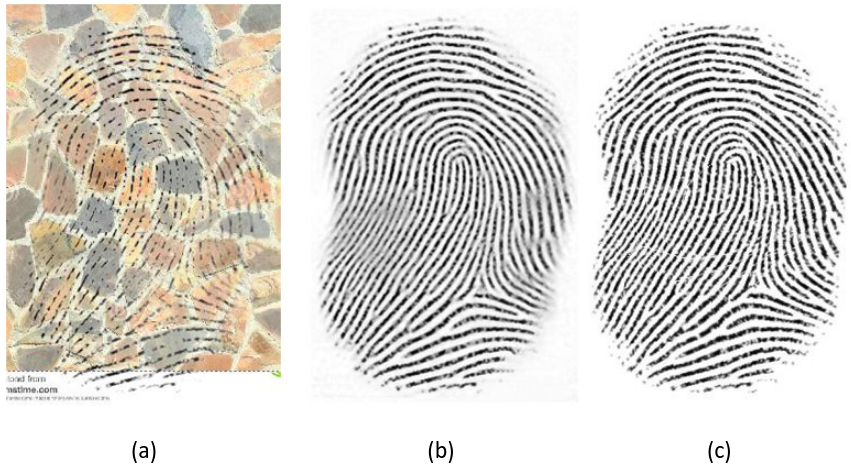}
	\caption{Moderate loss in fingerprint, (a) Original, (b) U-Finger (c) Ground truth.}
	\label{fig3}       
\end{figure}

\begin{figure}[!tbh]
	\centering
	\includegraphics[width=95mm]{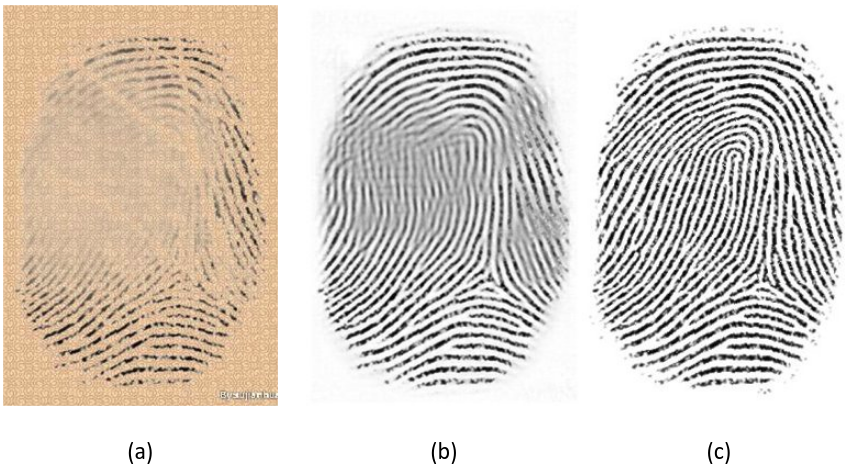}
	\caption{Severe loss in fingerprint, (a) Original (b) U-Finger (c) Ground truth.}
	\label{fig4}     
\end{figure}

Fig. 3 displays an example of U-Finger inpainting the moderate loss of fingerprint textures. Note that the inpaining is in a ``blind" setting, i.e., the model does not know the lost regions in advance. Even when the fingerprint loss gets more severe, e.g., as in Fig. 4, U-Finger still produces impressive recovery.
\\
\\

\section{Conclusion and Future Work}
We have proposed the U-Finger model for finger image restoration. The multi-scale nested architecture with up-sampling and down-sampling modules proves to achieve compelling balance between preserving fine texture and suppressing artifacts. The usage of dilated convolutions and the removal of padding have further boosted the performance. Our future work will include training with alternative loss functions \cite{liu2018improved}, as well as trying more densely connected modules.

\end{document}